\documentclass[10pt]{scrartcl}

\usepackage{makeidx}
\usepackage{graphicx}
\usepackage{amssymb}
\usepackage{amsmath}
\usepackage{amscd}
\usepackage{tikz}
\usetikzlibrary{matrix,arrows}
\usepackage{multicol}
\newtheorem{theorem}{Theorem}

\newcommand{\qed}{\hspace*{\fill} $\blacksquare$}
\newcommand{\proof}{Proof:\ }

\newcommand{\commentout}[1]{}

\newcommand{\R}{\mathbb{R}}                    % reals
\newcommand{\abs}[1]{\mathop{\left\lvert #1 \right\rvert}} 
\newcommand{\args}[1]{\mathop{\left( #1 \right)}} 

\newcommand{\norm}[1]{\mathop{\left\lVert #1 \right\rVert}}
\newcommand{\cbrace}[1]{\mathop{\left\{ #1 \right\}}}
\newcommand{\bracket}[1]{\mathop{\left[ #1 \right]}}

\newcommand{\cbraceS}[2]{\mathop{\left\{ #1 \right\}#2}}

\DeclareMathOperator{\id}{id}                  % identity function
\DeclareMathOperator{\sgn}{sgn}                % signum function

\renewcommand{\S}[1]{{\mathcal{#1}}}           % arbitrary sets
\def\vec#1{\mathchoice{\mbox{\boldmath$\displaystyle#1$}}
{\mbox{\boldmath$\textstyle#1$}}
{\mbox{\boldmath$\scriptstyle#1$}}
{\mbox{\boldmath$\scriptscriptstyle#1$}}}

\renewenvironment{cases}{%
\left\{\begin{array}{c@{\quad : \quad}l}}%
{%
\end{array}\right.}

\newcounter{algorithm_counter}
{
\\[-1.5ex]
\rule{\textwidth}{\arrayrulewidth}
\end{footnotesize}
\end{minipage}
\setlength{\parindent}{\parindent}
}

{
\\[-1.5ex]
\rule{\textwidth}{\arrayrulewidth}
\end{footnotesize}
\end{minipage}
\setlength{\parindent}{\parindent}
}
\begin{document}

\title{Sublinear Models for Graphs}

\author{Brijnesh J.~Jain \\
       Technische Universit\"at Berlin\\
       Berlin, Germany\\
       e-mail: brijnesh.jain@gmail.com}
            
\date{}
\maketitle

\begin{abstract} 
This contribution extends linear models for feature vectors to  sublinear models for graphs and analyzes their properties. The results are (i) a geometric interpretation of sublinear classifiers, (ii) a generic learning rule based on the principle of empirical risk minimization, (iii) a convergence theorem for the margin perceptron in the sublinearly separable case, and (iv) the VC-dimension of sublinear functions. Empirical results on graph data show that sublinear models on graphs have similar properties as linear models for feature vectors.
\end{abstract} 
\section{Introduction}

Linear classifiers and the k-nearest neighbor rule are two simple but powerful and well-investigated methods. A linear classifier is computationally efficient, makes strong assumptions about structure and yields stable but possibly inaccurate predictions \cite{Hastie2001}. In addition, linear classifiers, for example in the guise of the perceptron, form an elementary building block for more powerful methods such as deep learning \cite{Hinton2006} and other neural network architectures \cite{Haykin1994}. In contrast, the k-nearest neighbor rule is computationally less efficient, makes mild assumptions about structure and its predictions are often accurate but can be unstable \cite{Hastie2001}. 

While the k-nearest neighbor rule is applicable to any distance space, linear classifiers are confined to inner product spaces. The impact of linear classifiers in pattern recognition gives rise to the following questions: (1) How can we extend linear classifiers to domains other than vector spaces? (2) Suppose we can answer the first question. How can we efficiently learn such a classifier and what are its basic properties?

There are two main research directions related to the wider context of both questions. The first approach embeds patterns into a vector space and then applies linear classifiers \cite{Pekalska2005, Riesen2007,Livi2014}. The other approach defines a suitable positive semi-definite kernel for support vector learning \cite{Gaertner2003, Haussler1999}. None of these directions provides an answer to either of both questions. 

In this contribution, we extend linear models from feature vectors to graphs. Given a suitable graph similarity function, we extend linear functions to functions of the form $f(X) = \text{sim}(W, X) + b$, where $X$ is an input graph, $W$ is a weight graph, and $b \in \R$ is the bias. We show that functions $f(X)$ on graphs correspond to piecewise linear function in some Euclidean vector space and provide a geometrical interpretation of the decision surface separating the class regions. Since those piecewise linear functions are sublinear, we refer to $f(X)$ as a sublinear function on graphs. We derive a generic update rule for learning the weight graph $W$ and present a perceptron convergence theorem for a separable sample of graphs. Using the geometrical insight, we derive the VC dimension of sublinear functions and discuss the issue of structural risk minimization for model selection. In an empirical study, we show that the margin perceptron algorithm for graphs has similar properties as known from its vectorial counterpart. 

\section{Sublinear Functions on Graphs}

This section introduces sublinear functions  and the graph domain they are defined on.

\subsection{The Space of Attributed Graphs}

Let $\mathbb{A} = \R^d$ be a set of node and edges attributes. We consider graphs of the form $X = (\S{V}, \S{E}, \S{A})$, where $\S{V}$ represents a set of vertices, $\S{E}$ a set of edges, and $\S{A}\subseteq\mathbb{A}$ a subset of attributes of the nodes and edges.  Node attributes take the form $\vec{x}_{ii} \in \S{A}$ for each node $i \in \S{V}$ and edges attributes are given by $\vec{x}_{ij} \in \S{A}$ for  each edge $(i,j) \in \S{E}$. By $\S{X_G}$ we denote the space of all graphs with attributes from $\mathbb{A}$.

\subsection{Sublinear Dot Product}

We equip the space $\S{X_G}$ with a similarity function derived from the dot product of vectors. Given two graphs $X$ and $Y$, the desired similarity function aims at finding one-to-one correspondences between nodes of graph $X$ and nodes of graph $Y$ such that the sum of corresponding node and induced edge similarities is maximized.

We encode node correspondences by a binary match-matrix $\vec{M} = (m_{ij})  \in \cbraceS{0,1}{^{m \times n}}$, where $m$ and $n$ are the number of nodes of $X$ and $Y$, respectively. If node $i$ of $X$ is assigned to node $r$ of $Y$, then $m_{ir} = 1$, and $m_{ir} = 0$ otherwise. In addition, we demand that
\begin{align}
\label{eq:constr1}
\sum_{i} m_{ir}& \leq 1 \quad \forall r\\
\label{eq:constr2}
\sum_{r} m_{ir} &\leq 1 \quad \forall i\\
\label{eq:constr3}
\sum_{ir} m_{ir} &= \min(n, m).
\end{align}
The constraints ask for one-to-one correspondences that can not be extended to larger subset of nodes. The first constraint enforces that any node of $Y$ corresponds to at most one node from $X$. The second constraint enforces that any node of $X$ corresponds to at most one node of $Y$. Finally, the last constraint  requires that each node of the smaller graph corresponds to exactly one node of the larger graph.  By  $\S{M}^{n \times m}$ we denote the set of all binary  $n \times m$-match-matrices satisfying \eqref{eq:constr1}-\eqref{eq:constr3}.

The second concept for defining our desired similarity function are node and edge similarities. For this, we use the dot product defined on the attribute set $\mathbb{A}$. Then each match-matrix $M \in \S{M}^{n \times m}$ gives rise to a kernel on $X$ and $Y$ 
\[
k_M(X, Y) = \sum_{i,j,r,s} m_{ir}m_{js} \vec{x}_{ij}^T\vec{y}_{rs},
\]
 where $\vec{x_{ij}}^T\vec{y}_{rs}$ denotes the dot product between attribute vectors $\vec{x}_{ij}$ of $X$ and $\vec{y}_{rs}$ of $Y$. By maximizing the kernel $k_M$ over all match matrices, we obtain the sublinear dot product 
\begin{align}\label{eq:sdp}
X \cdot Y = \max_{\vec{M} \in \S{M}^{n \times m}} k_M(X, Y).
\end{align}
The sublinear dot product extends the dot product from vectors to graphs. It is straightforward to verify that the function $f_Y(X) = X \cdot Y$  as a pointwise maximizer of dot products is sublinear, that is convex and positively homogeneous. For this reason, we call $X\cdot Y$ sublinear. Though the sublinear dot product is not linear, it shares similar geometrical properties and generalizes the concept of maximum common subgraph \cite{Jain2009}. It can be reduced to a special case of the graph-edit distance and is widely used in different guises as a common choice of proximity measure for graphs \cite{Caetano2007,Cour2006,Gold1996,Umeyama1988,Wyk2002}.

\subsection{Sublinear Functions}
A sublinear function on graphs is of the form
\begin{align}\label{eq:discriminant}
f(X) = W \cdot X + b,
\end{align}
where $W$ is the weight graph and $b \in \R$ is the bias. As usual, we assign graph $X$ to the positive class if $f(X)\geq 0$, and to the negative class otherwise. 
Then equation $f(X) = 0$ defines a decision surface 
\[
\S{H}_f = \cbrace{X \in \S{X_G}\,:\, f(X) = 0}\subseteq \S{X_G}
\]
that separates the graph space $\S{X_G}$ into two class regions. 

\commentout{
The class regions $\S{R}_+$ for the positive and $\S{R}_-$ for the negative class are of the form
\begin{align}
\S{R}_+ &= \cbrace{X \in \S{X_G} \,:\, f(X; W, b) \geq 0}\\
\S{R}_- &= \cbrace{X \in \S{X_G} \,:\, f(X; W, b) < 0}.
\end{align}
Using sublinear functions, we can classify graphs in the usual way: assign graph $X$ to class  $y = +1$ if $f(X)\geq 0$, and to class $y = -1$ otherwise. 
}

\section{Graph Orbifolds}

For studying sublinear functions on graphs, we need a more convenient representation of the graph space. We suggest to represent attributed graphs as points of some graph orbifold as proposed by \cite{Jain2009,Jain2012}. 

\subsection{Graphs as Points in Orbifolds}

For the sake of mathematical convenience, we make the following assumptions: (1) attributes $\vec{x}_{ij}\in \S{A}$ of edges $(i,j)\in \S{E}$ are non-zero; and (2) all graphs are of order $n$, where  $n$ is chosen to be sufficiently large. 

Assumption (1) allows us to restrict to complete graphs, where non-edges are represented by edges with attribute zero.\footnote{If the zero-vector is a valid edge-attribute, an extra dimension serving as an edge-flag can be introduced to ensure that edges have non-zero attribute.}  
Assumption (2) can be satisfied by adding isolated vertices with attribute zero to graphs of order less than $n$ to obtain a graph of order $n$.
Both assumptions are purely technical without computational and limiting impact in practice \cite{Jain2009}. Both assumptions are purely technical without computational and limiting impact in practice. In a practical setting, we neither extend graphs to a larger order nor care about the maximum order $n$ otherwise. See  \cite{Jain2009} for more details.

Under these assumptions, a graph $X$ is completely specified by a matrix representation  $\vec{X} = (\vec{x}_{ij})$, where the elements $\vec{x}_{ij}$ are the node and edges attributes. By $\S{X} = \mathbb{A}^{n \times n}$, we denote the Euclidean space of ($n \times n$)-matrices with elements from $\mathbb{A}$.

The particular form of a matrix representation of a graph depends on how we order its nodes. Permuting the order of the nodes of a graph may result in a different matrix representation. By permuting the nodes in all possible ways, we obtain the equivalence class $[\vec{X}]$ of all matrix representations of $X$. 

Mathematically, we can describe the equivalence class $[\vec{X}]$ of matrix representations as the orbit of $\vec{X}$ under the action of a permutation group $\S{G}$ on $\S{X}$. Let $\S{S}_n$ be the symmetric group consisting of all permutations of $n$ elements. Each permutation $\phi \in \S{S}_n$ can be regarded as a bijective mapping $\phi:\S{V} \to \S{V}$
that reorders the nodes. The permutation $\phi$ induces a mapping on the set $\S{X}$ of matrix representations:
\[
g_\phi: \S{X} \rightarrow \S{X}, \quad \vec{X} \mapsto g_\phi(\vec{X}),
\]
where $g_\phi(\vec{X})$ is the matrix representation obtained from $\vec{X}$ by reordering the vertices via the permutation $\phi$. Then $\S{G} = \cbrace{g_{\phi}\,:\, \phi \in \S{S}_n}$ is a finite group acting on $\S{X}$. For a matrix representation $\vec{X} \in \S{X}$, the orbit of $\vec{X}$ is the set defined by
\[
\bracket{\vec{X}} = \cbrace{g(\vec{X}) \,:\, g \in \S{G}}. 
\]
In the following, we use vector- rather than matrix-based notations by stacking the columns of a matrix $\vec{X}$ to its vectorized replica $\vec{x}$.  Then $\vec{x}$ is a vector representation of graph $X$, if $\pi$ projects $\vec{x}$ to $X$. We identify orbits $\bracket{\vec{x}}$ with graphs $X$ and occasionally write $\vec{x} \in X$, if vector $\vec{x}$ projects to (represents) graph $X$.

\subsection{Sublinear Dot Product}

In a graph orbifold $\S{X_G}$, the structural dot product defined in \eqref{eq:sdp} is of the form
\begin{align*}\label{eq:sdp2}
X \cdot Y &= \max \cbrace{ \vec{x}^T\vec{y} \,:\, \vec{x} \in X, \vec{y} \in Y}\\
&= \max \cbrace{ \vec{x}^T\vec{y} \,:\, \vec{x} \in X}\\
&=\max \cbrace{ \vec{x}^T\vec{y} \,:\, \vec{y} \in Y},
\end{align*}
where $\vec{y} \in Y$ in the second and $\vec{x} \in X$ in the third line are arbitrarily chosen representations. Equality follows by symmetry and the properties of the group action \cite{Jain2009}. Of particular interest are representations $\vec{x}\in X$ and $\vec{y} \in Y$ that attain the maximum, that is $X \cdot Y = \vec{x}^T\vec{y}$. In this case, we say $\vec{x}$ and $\vec{y}$ are optimally aligned.

\section{Geometrical Interpretation}

Instead of studying geometrical properties of functions and sets in the graph space $\S{X_G}$, it is more convenient to study lifted versions of those objects in the ambient Euclidean space $\S{X}$. To this end, we first define the concept of lifting. Then we apply this concept to sublinear functions $f(X)$ and decision surfaces $\S{H}_f$. Finally, we present a geometrical interpretation.

\subsection{Lifting Functions and Sets}

Suppose that $f:\S{X_G} \rightarrow \R$ is a real-valued function defined on graphs. Then the lift $f^\ell$ of $f$ is the unique function defined on the ambient Euclidean vector space $\S{X}$ such that the following diagram commutes
\begin{center}
\begin{tikzpicture}[description/.style={fill=white,inner sep=2pt}]
\matrix (m) [matrix of math nodes, row sep=3em, column sep=2.5em, text height=1.5ex, text depth=0.25ex]
{ \S{X} & & \S{X_G} \\
& \R & \\ };
\path[->, font=\small]
(m-1-1) edge node[description] {$ \pi $} (m-1-3)
edge node[description] {$ f^\ell $} (m-2-2)
(m-1-3) edge node[description] {$ f $} (m-2-2);
\end{tikzpicture}
\end{center}
Thus, we have have $f^\ell(\vec{x})= f(\pi(\vec{x}))$ showing that the lift $f^\ell$ is constant on an orbit $[\vec{x}]$, that is on all vectors $\vec{x}$ that represent the same graph $X = \pi(\vec{x})$. In a similar manner, we can define the lift of a subset $\S{U_G} \subseteq \S{X_G}$ by 
\[
\S{U}_{\S{G}}^\ell = \cbrace{\vec{x} \in \S{X} \,:\, \pi(\vec{x}) \in \S{U_G}}.
\]
Thus, the lift of a set $\S{U_G}$ of graphs is the set of all vectors that represent a graph from $\S{U_G}$.

\subsection{Lifted sublinear Functions}

The lift of $f(X) = W\cdot X +b$ is a unique function defined by
\begin{align}\label{eq:lift-sublinear-1}
f^\ell(\vec{x}) =  \max_{\vec{w} \in W} \vec{w}^T\vec{x} + b,
\end{align}
where $\vec{x} \in X$.  Equation \eqref{eq:lift-sublinear-1} shows that the function $f(X)$ on graphs can be viewed as a pointwise maximizer of linear functions $h(\vec{x}) = \vec{w}^T\vec{x} + b$ in the ambient Euclidean space $\S{X}$, parametrized by the different representations $\vec{w}$ of weight graph $W$.
From eqn.\ \eqref{eq:lift-sublinear-1} follows that $f^\ell$ is a piecewise linear function. Thus, sublinear functions on graphs correspond to piecewise linear functions in the ambient Euclidean space.

The lift of the decision surface $\S{H}_f$ defined by the equation $f(X) = 0$ is of the form
\[
\S{H}_f^\ell = \cbrace{\vec{x} \in \S{X} \,:\, f(\pi(\vec{x})) = 0}.
\]
Since $f(\pi(\vec{x})) = f^\ell(\vec{x})$, we see that the lifted surface $\S{H}_f^\ell$ coincides with the surface $\S{H}_{f^\ell}$ defined by the lifted equation $f^\ell(\vec{x}) = 0$.

\subsection{Truncation}
By the properties of the structural dot product, we can express a sublinear function as
\[
f(X) =  \max_{\vec{x} \in X} \;\vec{w}^T\vec{x} + b,
\]
where $\vec{w}$ is an arbitrarily chosen representation of weight graph $W$. To evaluate $f$ at $X$, we select a representation $\vec{x}$ of $X$ that is closest to $\vec{w}$ and then compute the linear function $\vec{w}^T\vec{x} + b$. 

Suppose that $\phi:\S{X_G} \rightarrow \S{X}$ is a map satisfying the following two properties for all graphs $X \in \S{X_G}$:
\begin{enumerate}
\item $\pi \circ \phi = \id$
\item $W \cdot X = \vec{w}^T\phi(X)$
\end{enumerate}
The first property states that $\phi$ maps each graph to one of its representations. Mappings with this property are called cross sections. The second property states that the cross section $\phi$ selects for each graph a representation that is closest to $\vec{w} \in W$. We call $\phi$ a fundamental cross section along $\vec{w}$. 

Note that $\phi$ is not uniquely determined, because there are graphs that may have several different representations satisfying both properties of a fundamental cross section. As an immediate consequence, the image $\phi(\S{X_G})$ is also not unique, however, the closure of $\phi(\S{X_G})$ is well-defined and of the form
\[
\S{D}_{\vec{w}} = \cbrace{\vec{x} \in \S{X} \,:\, \vec{w}^T\vec{x} \geq \widetilde{\vec{w}}^T\vec{x}, \, \widetilde{\vec{w}} \in W}.
\]
We call the closure $\S{D}_{\vec{w}}$ the Dirichlet (fundamental) domain of $\vec{w}$. A Dirichlet domain is a convex polyhedral cone \cite{Jain2012} with the following properties: 
\begin{enumerate}
\item $\vec{x} \in \S{D}_{\vec{w}}$ iff $(\vec{x}, \vec{w})$ is an optimal alignment
\item $\pi(\S{D}_{\vec{w}}) = \S{X}_G$
\item $\pi$ is injective on the interior of $\S{D}_{\vec{w}}$
\end{enumerate}
From these properties we can draw the following conclusions: (i) Since the natural projection $\pi$ restricted to $\S{D}_{\vec{w}}$ is surjective, each graph $X$ has at least one representation $\vec{x}$ in $\S{D}_{\vec{w}}$. (ii) Since the natural projection $\pi$ is injective on the interior of $\S{D}_{\vec{w}}$, different representations $\vec{x}$ and $\vec{x}'$ of the same graph lie on the border of  $\S{D}_{\vec{w}}$. This implies that graphs with multiple representations in $\S{D}_{\vec{w}}$ form a set of Lebesgue measure zero. 

From the foregoing discussion follows that studying sublinear functions $f$ can be reduced to the study of truncated lifts  $f^t$ defined by the restriction of the lift $f^\ell$ to some Dirichlet domain $\S{D}_{\vec{w}}$. The truncated lift $f^t$ is a linear function 
\[
f^t(\vec{x}) = \vec{w}^T\vec{x} + b
\]
locally defined on $\S{D}_{\vec{w}}$.

\subsection{Geometrical Interpretation}

Linear functions $h(\vec{x}) = \vec{w}^T\vec{x} + b$ on feature vectors give rise to the following geometrical  interpretation:  
\begin{enumerate}
\item the decision surface $\S{H}_{\vec{w},b}$ defined by $h(\vec{x}) = 0$ is a hyperplane; 
\item the weight vector $\vec{w}$ is normal to the hyperplane $\S{H}_{\vec{w},b}$; 
\item $b/\norm{\vec{w}}$ is the distance of the hyperplane $\S{H}_{\vec{w},b}$ from the origin; and
\item $h(\vec{x})/\norm{\vec{w}}$ is an algebraic measure of the distance of $\vec{x}$ from the hyperplane $\S{H}_{\vec{w},b}$. 
\end{enumerate}
We show that  these four geometrical properties carry over to sublinear functions $f(X) = W\cdot X + b$ to a certain extent.

\paragraph*{Property 1: } 
The decision surface $\S{H}_f$ of a sublinear function $f$ in the graph space $\S{X_G}$ corresponds to a piecewise linear decision surface in the ambient Euclidean space $\S{X_G}$.
The lifted decision surface is of the form
\[
\S{H}_f^\ell = \cbrace{\vec{x} \in \S{X} \,:\, f^\ell(\vec{x}) = \max_{\vec{w} \in W} \vec{w}^T\vec{x} + b = 0}.
\]
Since the lift $f^\ell$ is piecewise linear, the decision surface $\S{H}_f^\ell$ is piecewise linear. We can express $\S{H}_f^\ell$ as the union of hyperplane segments
\[
\S{H}_f^\ell = \bigcup_{\vec{w} \in W} \S{H}_{\vec{w}},
\]
where each hyperplane segment $\S{H}_{\vec{w}}$ is defined by
\[
\S{H}_{\vec{w}} = \cbrace{\vec{x} \in \S{X} \,:\, f^\ell(\vec{x}) = \vec{w}^T\vec{x} + b = 0}.
\]
The decision surface of the truncated lift $f^t$ on $\S{D}_{\vec{w}}$ is the intersection 
\[
\S{D}_{\vec{w}} \cap \S{H}_f^\ell
\]
and coincides with the hyperplane segment $\S{H}_{\vec{w}}$

\paragraph*{Property 2:}
Each representation $\vec{w}$ of the weight graph $W$ is normal to the hyperplane segment $\S{H}_{\vec{w}}$.  

\paragraph*{Property 3:}
The distance of the decision surface $\S{H}_f$ from the zero graph is given by
\[
d(\S{H}_f, 0) = \frac{b}{\sqrt{W \cdot W}}.
\]
Observe that the distance of each linear hyperplane segment  $\S{H}_{\vec{w}}$ from the origin is constant with value
\[
\frac{b}{\norm{\vec{w}}} =  \frac{b}{\sqrt{W \cdot W}},
\]
because the norm $\norm{\vec{w}}$ is invariant under permutations. 

\paragraph*{Property 4:} 
The value 
\[
\frac{f(X)}{\sqrt{W \cdot W}} \leq d(X, \S{H}_f)
\]
is a lower bound of the distance $d(X, \S{H}_f)$ of $X$ from the hyperplane $\S{H}_f$. Equality holds, if there is a representation $\vec{x}\in X$ and a representation $\vec{w} \in W$ such that $\vec{x}$ has an orthogonal projection onto the hyperplane segment $\S{H}_{\vec{w}}$.

\section{Learning}

Learning typically amounts in minimizing a differentiable risk function using local gradient information. The concept of derivative, however, is unknown for functions on graphs. We solve this problem by first lifting the risk function on graphs to the Euclidean space and then applying stochastic subgradient method for minimizing the lifted risk.

\subsection{The Learning Problem}
The goal of learning consists in finding a weight graph $W$ and bias $b$ such that the sublinear discriminant $f(X) = W\cdot X + b$ minimizes the expected risk
\begin{align}\label{eq:expected-risk}
E(f) = \int_{\S{X_G}} L(f(X), y)\: dP(X, y),
\end{align}
where $L(\hat{y}, y)$ is a differentiable loss function that measures the cost of predicting class $\hat{y}$ when the actual class is $y$ and $P(X, y)$ is the joint probability distribution on $\S{X_G} \times \S{Y}$. 
Since the distribution $P(X, y)$ is usually unknown, the expected risk $E(f)$ can not be computed directly. Instead, we approximate the expected risk $E(f)$ by minimizing the empirical risk
\[
E_N(f) = \frac{1}{N}\sum_{i=1}^N L(f(X_i), y_i)
\]
on the basis of $N$ training examples $(X_i, y_i) \in \S{X_G} \times \S{Y}$.

\subsection{The Lifted Risk}
The lift of the expected risk is of the form
\begin{align}\label{eq:liftedrisk}
E_N^\ell \args{f}= \frac{1}{N}\sum_{i=1}^N \,L\!\args{\max_{\vec{x}_i\in X_i} \vec{w}^T\vec{x}_i + b, y_i}
\end{align}
From eqn.\ \eqref{eq:liftedrisk} directly follows
\[
E_N(f) = E_N^\ell\!\args{f} = E_N\!\args{f^\ell}.
\]
Since we minimize the lifted risk $E_N^\ell(f)$ as a function of $(\vec{w}, b)$, it is more convenient to express the lift $f^\ell$ as
\[
f^\ell(\vec{w}, b | X) = \max_{\vec{x} \in X} \vec{w}^T\vec{x} + b.
\]
The lift $f^\ell$ is convex as a function of $(\vec{w}, b)$ and non-differentiable on a subset with probability zero. Since the loss $L$ is convex and differentiable, the same holds for the composition  $L(f^\ell(\vec{w}, b | X), y)$.

\subsection{Stochastic Subgradient Method}
Our goal is to minimize the lifted risk $E_N^\ell(f)$  as a function of $(\vec{w}, b)$ in an iterative fashion. Each iteration $t$ consists of choosing a training example $(X_t, y_t)$ at random, and updating the weights $\vec{w}$ and the bias $b$ according to the following learning rule
\begin{align*}
\vec{w}_{t+1} &= \vec{w}_t + \eta_t\cdot\partial_{\vec{w}_t}\\
b_{t+1} &= b_t + \eta_t\cdot\partial_{b_t},
\end{align*}
where $0 < \eta_t$ is the learning rate and $\partial_t = (\partial_{\vec{w}_t}, \partial_{b_t})$ is a suitable direction along which we move the point $(\vec{w}_t, b_t)$ in order to minimize $E_N^\ell(f)$.
Given the loss at iteration $t$ 
\begin{align*}
E_t(\vec{w}, b) = L(\hat{y}_t, y_t) = L\args{f^\ell\args{\vec{w}_t, b_t | X_t}, \:y_t},
\end{align*}
we define the update direction $\partial_t$ by 
\begin{align*}
\partial_{\vec{w}_t} &= -\nabla L(\hat{y}_t, y_t)\cdot\vec{x}_t \\
\partial_{b_t} &= -\nabla L(\hat{y}_t, y_t),
\end{align*}
where $\vec{x}_t$ is a representation of $X_t$ optimally aligned with $\vec{w}_t$, and $\nabla L(\hat{y}_t, y_t)$ is the derivative of $L$ as a function of $\hat{y}_t$.

At differentiable points $(\vec{w}_t, b_t)$ the direction $\partial_t$ is exactly the opposite direction of the gradient of $E_t(\vec{w}, b)$. In this case, the update rule of learning a sublinear classifier on graphs coincides with its counterpart in vector spaces. At non-differentiable points $(\vec{w}_t, b_t)$ the update direction $\partial_t$ satisfies the inequality
\begin{align*}\label{eq:subgradient}
E_t(\vec{w}, b) \geq E_t(\vec{w}_t, b_t) + \partial_t^T\args{(\vec{w},b) - (\vec{w}_t, b_t)}
\end{align*}
for all $(\vec{w}, b)$. Any vector satisfying the above inequality is a subgradient. If $E_t$ is convex, a subgradient exists at each point $(\vec{w}, b)$. At differentiable points the subgradient of  $E_t$ is unique and coincides with its gradients.

\medskip

\noindent
\textbf{Example (Margin Perceptron):}
Suppose that $\vec{x}$ and $\vec{w}$ are optimally aligned representations. The hinge loss of the margin perceptron is of the form
\[
L(\vec{w}^T\vec{x} + b, y) = \max\cbrace{0, \lambda-y\cdot(\vec{w}^T\vec{x} + b)},
\]
where $\lambda$ is the margin. The partial direction $\partial_{\vec{w}}$ takes the form
\[
\partial_{\vec{w}} = 
\begin{cases}
-y \cdot \vec{x}& \text{if } y\cdot \args{\vec{w}^T\vec{x}+b} \leq \lambda  \\
0 & \text{otherwise}
\end{cases}.
\]
Setting the margin to $\lambda = 0$ yields the perceptron algorithm as a special case. 

\subsection{Margin Perceptron Convergence Theorem}
The perceptron convergence theorem states that the perceptron algorithm with constant learning rate finds a separating hyperplane, whenever the training patterns are linearly separable. We provide a weaker convergence result for the margin perceptron algorithm in the graph domain.

Suppose that  $\S{S} \subseteq \S{X_G} \times \S{Y}$ is a sample consisting of $N$ graphs $X \in \S{X_G}$ with corresponding class labels $y \in \S{Y}$. The sample is sublinearly separable if there is a weight graph $W$ and bias $b$ such that 
\[
\sgn\args{W \cdot X + b} = y
\]
for all $(X, y) \in \S{S}$. We say, $\S{S}$ is sublinearly separable with margin $\xi > 0$ if 
\[
\min_{(X,y) \in \S{S}} \;y \args{W \cdot X+ b} \geq \xi.
\]
Now we can state the margin perceptron convergence theorem. 
\begin{theorem}
Suppose that $\S{S}\subseteq \S{X_G} \times \S{Y}$ is sublinearly separable with margin $\xi > 0$. Then the margin perceptron algorithm with fixed learning rate $\eta$ and margin-parameter $\lambda \leq \xi$ converges to a solution $(W, b)$ that correctly separates the sample $\S{S}$ after a finite number of update steps, provided the learning rate is chosen sufficiently small. 
\end{theorem}

\proof 
Suppose that $\abs{\S{S}} = N$. Since sublinear functions are convex, the sublinear dot product is convex.  In addition, the composition of convex functions is convex. Hence, for each example $(X_i, y_i) \in \S{S}$ the function 
\[
F_i(W, b) = L(W \cdot X_i + b, y_i)
\]
is real-valued and convex. Then there is a positive scalar $C_i$ that bounds the subdifferential of $F_i$ at $X_i$. Suppose that
\[
C = \max_{i = {1, \ldots, N}} C_i.
\]
Let 
\[
F(W, b) = \sum_{i=1}^N F_i(W, b)
\]
the risk without averaging operation, that is $F = N \cdot E_N$, where $E_N$ is the empirical risk. Since the sample $\S{S}$ is sublinearly separable by assumption, the minimum $F_*$ of $F(W, b)$ is zero.  From \cite{Nedic2001}, Prop.Ê 2.1. follows
\[
\lim_{t \to \infty} F(W_t, b_t) \leq F_* + \frac{\eta \cdot C^2}{2} = \frac{\eta \cdot C^2}{2},
\] 
where $\eta$ is the learning rate. Choosing 
\[
\eta \leq \frac{\xi}{C^2}
\]
gives
\[
\lim_{t \to \infty} F(W_t, b_t) \leq \frac{\xi}{2}.
\] 
Since $\xi > 0$, this implies that there is a $t_0$ such that 
\[
F_t(W_t, b_t) < \xi
\]
for all $t \geq t_0$. Here, $F_t$ refers to the example $(X_t, y_t) \in \S{S}$ presented at time $t$. From this follows that all training examples are classified correctly after a finite number of update steps, provided that $\lambda \leq \xi$.
\qed

\subsection{VC-Dimension}
Via lifting and truncating, we can show that sublinear functions on graphs correspond to linear functions restricted to some Dirichlet domain in the ambient Euclidean space $\S{X}$. From this directly follows that the  VC-dimension of sublinear functions is equal to $dim(\S{X}) + 1$. 

In what follows, we use the generic term \emph{item} to refer to either nodes or edges of some graph. Since graphs may have different number of items, we can control the capacity of sublinear functions by the number of items of $W$. By strictly applying the update rule, the number of nodes of $W$ after learning will be equal to the number of nodes of the largest graph in the training set. A similar statement about the number of edges is not possible. If many graphs are smaller than the largest graph and if the classification problem is simple, some substructures of the weight graph $W$ may rarely undergo an update. In this case, the classifier is too complex and we can shrink the number of items of $W$. Hopefully only the relevant parts of the training graphs are aligned against $W$. Shrinking $W$ projects the problem to a lower-dimensional subspace.

In other cases, the classification problem is harder such that the size of the weight graph $W$ may not capture the distribution of the pattern graphs. Then the classifier is too simple and we may increase the number of items of $W$, even beyond the size of the largest graph in the training set. Due to the matching process, the training graphs will be embedded into a hopefully discriminative substructure of $W$. Growing $W$ projects the problem to a higher dimensional space.

Since we can control the VC-dimension of sublinear classifiers via the size of the weight graph, we can perform structural risk minimization for model selection. This issue is out of scope. Here, the number of nodes of $W$ is set to the number of nodes of the largest graph of the training set and no constraints are imposed on the number of edges.

\section{Experiments}
 
The goal of our empirical study is to assess the performance of the perceptron and margin perceptron algorithm for graphs.

\subsection{Data} 
We considered the coil, letter (low, medium, high), fingerprint, and grec dataset from the IAM graph database repository  \cite{Riesen2008}. All datasets are split into a predefined training, validation, and test set. 
For two-class problems, we selected subsets of the \emph{coil} data set. COIL$_{2,4}$ consists of graphs representing two types of ball-shaped fruits (index 2 \& 4). COIL$_{8,15}$  consists of graphs representing two types of cans (index 71 \& 93). 
All other datasets refer to multi-class problems. The \emph{letter} data sets consist of graphs representing distorted copies of 15 letter drawings from the Roman alphabet that consist of straight lines. The distortion levels are low, medium, and high. The \emph{fingerprints} data consist of graphs representing fingerprint images from four classes. The  \emph{grec} data  consists of graphs representing symbols from noisy versions of 22 architectural and electronic drawings. 

For further details of the datasets including a description of how graphs were generated from the images, we refer to  \cite{Riesen2008}.

\subsection{Algorithms} We compared both sublinear classifiers against the nearest neighbor method and the support vector machine with similarity kernel \cite{Bunke2011, Riesen2008}.  

\subsection{Experimental Protocol.}
We applied the graduated assignment algorithm \cite{Gold1996} for solving graph matching problems. For multi-class problems, we wrapped both sublinear classifiers into a one-against-all classifier. 

In a first step, we selected the parameters for the perceptron algorithm (learning rate $\eta$) and the large-margin perceptron (learning rate $\eta$ and margin $\lambda$). 
For each $\eta =  0.01$, $0.05$, $0.1$, $0.3$, $0.5$, $0.7$, $0.9$ we trained the perceptron algorithm on the training set and evaluated the learned model on the validation set. We conducted this experiment $10$ times and selected the learning rate $\eta_*$ which resulted in the best average performance on the validation set. 
Next, we trained the perceptron algorithm using $\eta_*$ on the training and validation set. Then we evaluated the learned model on the test set. We again conducted this experiment $10$ times and recorded the average accuracy, standard deviation, and the maximum accuracy. We applied the same experimental protocol for the large-margin algorithm. We adopted the optimal learning rate $\eta_*$ from the perceptron algorithm and selected the optimal margin from $\lambda = 0.01$, $0.05$, $0.075$, $0.1$, $0.125$, $0.15$, $0.2$.

Since the 1-NN method is deterministic, we performed a single run on each data set.  

\subsection{Results and Discussion}

\begin{table}
\begin{center}
\begin{footnotesize}
\begin{tabular}{|lc@{\quad}|c|c|cc|cc|}
\hline
\hline
data &\#(classes)&1-NN & sk-svm & \multicolumn{2}{c|}{perceptron}& \multicolumn{2}{|c|}{margin-perc}  \\
&& & & avg & max & avg & max  \\
\hline
&&&&&&& \\[-2ex]
COIL$_{2,4}$ &2 & 90.0 & --- & 95.0$^{\pm 0.0}\!\!\!\!\!\!$ & 95.0& 95.0$^{\pm 0.0}\!\!\!\!\!\!$ & 95.0 \\
COIL$_{71, 93}$ &2& 95.0 & ---&  96.0$^{\pm 5.2}\!\!\!\!\!\!$&100.0& 98.5$^{\pm 2.4}\!\!\!\!\!\!$&100.0\\ 
Letter L &15 & 95.6 & 99.2 & 94.5$^{\pm 0.7}\!\!\!\!\!\!$ & 96.0 & 95.5$^{\pm 0.3}\!\!\!\!\!\!$ & 95.7\\
Letter M &15 & 92.2 & 94.7 & 86.1$^{\pm 1.1}\!\!\!\!\!\!$ & 87.5 & 88.7$^{\pm 0.6}\!\!\!\!\!\!$ & 89.5\\
Letter H &15 & 83.9 & 92.8 & 80.7$^{\pm 1.1}\!\!\!\!\!\!$ & 82.5 & 84.1$^{\pm 0.5}\!\!\!\!\!\!$ & 84.8 \\
F'print &4 & 80.1 & 81.7 & 76.8$^{\pm 1.6}\!\!\!\!\!\!$ & 79.1 & 79.5$^{\pm 2.6}\!\!\!\!\!\!$ & 82.4 \\
GREC &22 & 97.5 & (92.2) & 96.3$^{\pm 0.5}\!\!\!\!\!\!$ & 97.0 & 97.5$^{\pm 0.6}\!\!\!\!\!\!$ & 98.1\\
\hline
\hline
\end{tabular}
\caption{Classification results. Results of perceptron and margin perceptron are averaged over $10$ runs. Standard deviation and maximum classification accuracy are shown for both sublinear models. Results are unknown for entries with '$-$'-sign.}
\label{tab:result}
\end{footnotesize}
\end{center}
\end{table}

Table \ref{tab:result} summarizes the results. Shown are the average and maximum classification accuracies as well as the standard deviations. Numbers in parentheses in the first line show the number of classes. Results of the SVM were taken form \cite{Bunke2011}. On GREC the SVM is not comparable to the other methods, because  \cite{Bunke2011} used a different version than the one originally published by \cite{Riesen2008}.

\paragraph*{Generalization performance.}
Both perceptron algorithms perform reasonable well compared to 1-NN and SVM. Results on the letter dataset with medium and high distortion level show that the assumption of sublinear separability is too strong and yields inaccurate predictions. On average, the margin perceptron performs slightly better than the standard perceptron algorithm. In case of COIL$_{71, 93}$, both perceptron algorithms show that they can learn a separating decision surface. These findings are similar to linear models in vector spaces: they are simple and powerful methods that  yield possibly inaccurate results.  Room for improvements is given by more extensively exploring the hyperparameters and controlling the VC-dimension via the size of the weight graph.

\paragraph*{Computational efficiency.}
The graph matching problem clearly predominates the computing time of all four classifiers. Both perceptron algorithms outperform the 1-NN and SVM approach. To classify a graph, the number of graph matching problems to be solved by a perceptron algorithm in a one-against-all setting is equal to the number of classes. In a two-class problem, only a single graph matching problem needs to be solved. In contrast, nearest  neighbor classifiers need to compare a graph against all training examples. The computing time of the SVM depends on the prototype selection algorithm for embedding the graphs into a vector space and by the number of support vectors. Results in \cite{Riesen2007} and follow-up publications by the same authors indicate that the SVM approach needs to compare each graph with $30\% - 60\%$ of all training examples.

\section{Conclusion}
This contribution generalizes linear classifiers to sublinear classifiers for graphs. To this end, we replaced the inner product of vectors by the sublinear dot product of graphs and lifted the learning problem to the ambient Euclidean space. The convergence theorem for the separable case carries over to the graph domain under mild assumptions. We can control the VC dimension via the size of the weight graph. Experiments show that perceptrons for graphs perform similarly as linear models for vectors. They are simple and efficient methods that yield possibly inaccurate predictions. Structural risk minimization for model selection and stability are prospects for further research. In addition, the stage is set for devising and analyzing further sublinear models as well as more sophisticated learning algorithms such as deep learning and other neural network architectures.

\end{document}